\documentclass{article}

\usepackage{arxiv}

\usepackage[utf8]{inputenc} 
\usepackage[T1]{fontenc}    
\usepackage{hyperref}       
\usepackage{url}            
\usepackage{booktabs}       
\usepackage{amsfonts}       
\usepackage{nicefrac}       
\usepackage{microtype}      
\usepackage{cleveref}       
\usepackage{lipsum}         
\usepackage{graphicx}
\usepackage[numbers]{natbib}
\usepackage{doi}

\title{A compendium of data sources for data science, machine learning, and artificial intelligence}

\date{This version (1.0): 10 September 2023}

\author{ \href{https://orcid.org/0000-0001-6846-6649}{\includegraphics[scale=0.06]{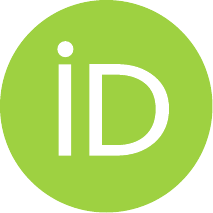}\hspace{1mm}Paul Bilokon} \\
	Imperial College London\\
	South Kensington Campus\\
	Exhibition Road\\
    London SW7 4UB\\
	\texttt{paul.bilokon@imperial.ac.uk} \\
	\And
	\href{https://orcid.org/0009-0006-9253-4502}{\includegraphics[scale=0.06]{orcid.pdf}\hspace{1mm}Oleksandr (Alex) Bilokon} \\
	Thalesians Marine Ltd\\
	3rd Floor, 120 Baker Street\\
	London W1U 6TU\\
	\texttt{alex@thalesiansmarine.com} \\
    \And
	\href{https://orcid.org/0000-0002-6068-4198}{\includegraphics[scale=0.06]{orcid.pdf}\hspace{1mm}Saeed Amen} \\
	Turnleaf Analytics Ltd\\
	59 Kensington Court\\
	London W8 5DG\\
	\texttt{saeed@turnleafanalytics.com} \\    
}

\renewcommand{\shorttitle}{A compendium of data sources}

\hypersetup{
pdftitle={A compendium of data sources for data science, machine learning, and artificial intelligence},
pdfsubject={q-bio.NC, q-bio.QM},
pdfauthor={Paul Bilokon, Oleksandr (Alex) Bilokon},
pdfkeywords={artificial intelligence, AI, machine learning, ML, data science, datasets, data, alternative data},
}

\begin{document}
\maketitle

\begin{abstract}
Recent advances in data science, machine learning, and artificial intelligence, such as the emergence of large language models, are leading to an increasing demand for data that can be processed by such models. While data sources are application-specific, and it is impossible to produce an exhaustive list of such data sources, it seems that a comprehensive, rather than complete, list would still benefit data scientists and machine learning experts of all levels of seniority. The goal of this publication is to provide just such an (inevitably incomplete) list --- or compendium --- of data sources across multiple areas of applications, including finance and economics, legal (laws and regulations), life sciences (medicine and drug discovery), news sentiment and social media, retail and ecommerce, satellite imagery, and shipping and logistics, and sports.
\end{abstract}

\keywords{artificial intelligence \and AI \and machine learning \and ML \and data science \and datasets \and data \and alternative data}

\section{Introduction}

Bearing in mind the recent advances in data science, machine learning, and artificial intelligence, such as the emergence of large language models~\cite{Euchner2023}, the availability of high-quality data is crucial for data scientists, machine learning, and artificial intelligence experts of all levels of seniority. Now that high-quality tools are available, access to data is a crucial enabling factor for research.

While data sources are application-specific, and it is impossible to produce an exhaustive list of such data sources, it seems that a comprehensive, rather than complete, list would still benefit the scientific and business communities. We intend to update this list as new data sources become available.

Many of the data sources listed here are what is known as \emph{alternative data}: data gathered outside of traditional sources such as company filings and broker research notes. It is not our goal to provide in this document instructions for making sense and extracting value from such data --- the reader may wish to consult~\cite{Denev2020}.

However, for the benefit of the reader, we will provide a few quick tips for working with these datasets. The technology of choice for dealing with data is the lingua franca of data science --- Python --- and its libraries, such as pandas, NumPy, and Matplotlib~\cite{McKinney2022}.

More advanced tools, such as boosting~\cite{Shapire2012}, deep learning~\cite{GoodBengCour16}, and ChatGPT~\cite{Euchner2023} can then be applied for high-quality data analysis, machine learning, and artificial intelligence. The process of artificial intelligence-assisted data analysis is quite involved, and usually requires practical expertise and an appropriate educational background, usually at the MSc or PhD level or equivalent.

The dataset may consist of historical data, which is static, or real-time data, which keeps arriving in real time. It may be relatively small in size or may constitute \emph{big data}~\cite{Novotny2019}, whose use may require special tools, such as specialised big data / high-frequency data databases (e.g. kdb+/q~\cite{Novotny2019}). If you intend to build a real-time production system utilising this data, you may need to build it in a programming language other than Python, such as C++~\cite{Stroustrup2022}.

The datasets may be commercial or noncommercial/free of charge. In each case, before using a dataset, make sure that you carefully examine the terms and conditions, such as licensing. Some vendors will provide \emph{application programming interfaces~(APIs)}~\cite{Jacobson2011}, which will enable you to easily access their product offering. You may have to use a general-purpose \emph{protocol}, such as REST~\cite{Ferreira2009} or WebSocket~\cite{Fette2011}, or a specialised one, such as the Financial Information Exchange~(FIX) protocol~\cite{FIXProtocol2023}, to access the dataset. Usually the relevant information is contained in the product documentation.

In some cases no API is provided, and the data may have to be extracted from websites (using libraries such as Selenium~\cite{Salunke2014} or Beautiful Soup~\cite{Richardson2007}), images or PDF files (using libraries such as Tesseract~OCR~\cite{Kay2007}). Before applying such \emph{scraping}~\cite{Mitchell2018} make sure that you have read the vendor's terms and conditions and confirm that the terms and conditions do indeed allow such usage. In each case it is generally a good idea to consult a legal professional before onboarding a dataset, especially a commercial dataset.

Sometimes you (or your organisation) may be your own best source of data --- in which case make sure that you log it carefully using an appropriate logging framework, database, or observability stack (Elastic\footnote{\url{https://www.elastic.co/}}, Grafana\footnote{\url{https://grafana.com/}}, Splunk\footnote{\url{https://www.splunk.com/}}, etc.).

You may wish to further automate your data science work using one or several of the business intelligence stacks (Alteryx\footnote{\url{https://www.alteryx.com/}}, Microsoft Power BI\footnote{\url{https://powerbi.microsoft.com/}}, Tableau Software~\footnote{\url{https://www.tableau.com/}}, etc.). Such tools can also make the data science, machine learning, and, in principle, artificial intelligence analysis more accessible to less technical users.

When selecting datasets for inclusion, we have of necessity been biased. We have therefore included first and foremost those datasets that have been used by us, our students, or other collaborators in our academic and/or commercial work. We do not guarantee the reliability of those datasets and the information is provided ``as is'' without any explicit or implicit warranty. The reader is reminded to check the relevant terms and conditions before using any dataset, including those mentioned here.

If a particular dataset is missing from this compendium and you would like to see it included in the following editions, please let us know. If there are mistakes and/or omissions, e.g. missing citations and/or URLs, please accept our apologies --- this is not intentional --- please also do let us know (ideally mentioning the suggested BibTeX, where appropriate).

\section{General resources}

Before we proceed to consider specific datasets, we will briefly mention some general-purpose tools that can help with dataset and machine learning technique search. It is never a good idea to reinvent the wheel, unless your intention is to replicate and validate existing results. Before starting your analysis, make sure that it hasn't already been done by someone else and does not appear in the literature. Therefore it's useful to perform a search for relevant academic papers and white papers before you commence your work. These are also good sources of up-to-date machine learning techniques.

\begin{enumerate}

    \item \emph{Google Dataset Search} is a search engine from Google that helps researchers locate online data that is freely available for use. The company launched the service on 5 September, 2018, and stated that the product was targeted at scientists and data journalists. The service was out of beta as of 23 January, 2020.

    URL: \url{https://datasetsearch.research.google.com/}
    
    \item \emph{Google Scholar} provides a simple way to broadly search for scholarly literature. It is a freely accessible web search engine that indexes the full text or metadata of scholarly literature across an array of publishing formats and disciplines. Released in beta in November 2004, the Google Scholar index includes peer-reviewed online academic journals and books, conference papers, theses and dissertations, preprints, abstracts, technical reports, and other scholarly literature, including court opinions and patents.

    URL: \url{https://scholar.google.com/}
    
    \item \emph{arXiv} is an open-access repository of electronic preprints and postprints (known as e-prints) approved for posting after moderation, but not peer reviewed. While such repositories provide early access to research, users should be aware that preprints have not been peer reviewed, and may not be of the same quality as peer reviewed papers in high-quality academic journals. Use at your own risk.

    URL: \url{https://arxiv.org/}
    
    \item \emph{medRxiv} is similar to arXiv but it distributes preprints specifically in health sciences.

    URL: \url{https://www.medrxiv.org/}
    
    \item \emph{Papers With Code} --- a free and open resource with machine learning papers, code, datasets, methods, and evaluation tables.

    URL: \url{https://paperswithcode.com/}

    \item \emph{GitHub} is a web-based version control and collaboration platform for software developers. It contains many software projects, many of which are public/open source, and can be searched.

    URL: \url{https://github.com/}

    \item \emph{GitLab} is a web-based version control and collaboration plarform for software developers, which also aims to be a comprehensive DevOps platform being delivered as a single application. It contains many software projects, many of which are public/open source, and can be searched.

    URL: \url{https://gitlab.com/}

    \item \emph{Kaggle} is the world's largest data science community with powerful tools and resources to help you achieve your data science goals. It features competitions, datasets, and tutorials.

    URL: \url{https://wwww.kaggle.com/}

    \item The \emph{Conference on Neural Information Processing Systems~(NeurIPS)} is a premier conference on machine learning and artificial intelligence. The conference was founded in 1987 and is now a multi-track interdisciplinary annual meeting that includes invited talks, demonstrations, symposia, and oral and poster presentations of refereed papers. Along with the conference is a professional exposition focusing on machine learning in practice, a series of tutorials, and topical workshops that provide a less formal setting for the exchange of ideas.

    URL: \url{https://nips.cc/}

    \item The \emph{International Conference on Machine Learning~(ICML)} is the premier gathering of professionals dedicated to the advancement of the branch of artificial intelligence known as machine learning. ICML is globally renowned for presenting and publishing cutting-edge research on all aspects of machine learning used in closely related areas like artificial intelligence, statistics, and data science, as well as important application areas such as machine vision, computational biology, speech recognition and robotics. ICML is one of the fastest growing artificial intelligence conferences in the world. Participants at ICML span a wide range of backgrounds, from academic and industrial researchers, to entrepreneurs and engineers, to graduate students and postdocs.

    URL: \url{https://icml.cc/}

    \item The \emph{International Conference on Learning Representations~(ICLR)} is the premier gathering of professionals dedicated to the advancedment of the branch of artificial intelligence called representation learning, but generally referred to as deep learning. ICLR is globally renowned for presenting and publishing cutting-edge research on all aspects of deep learning used in the fields of artificial intelligence, statistics, and data science, as well as important application areas such as machine vision, computational biology, speech recognition, text understanding, gaming, and robotics. Participants at ICLR span a wide range of backgrounds, from academic and industrial researchers, to entrepreneurs and engineers, to graduate students and postdocs.

    URL: \url{https://iclr.cc/}

    \item Wikipedia's \emph{List of datasets for machine-learning research} lists the datasets that are applied for machine learning research and have been cited in peer-reviewed academic journals. Datasets are an integral part of the field of machine learning. Major advances in this field can result from advances in learning algorithms (such as deep learning), computer hardware, and, less intuitively, the availability of high-quality training datasets. High-quality labelled training datasets for supervised and semi-supervised machine learning algorithms are usually difficult and expensive to produce because of the large amount of time needed to label the data. Although they do not need to be labelled, high-quality datasets for unsupervised learning can also be difficult and costly to produce. If you are developing machine learning algorithms, make sure that you compare them against these established benchmarks.

    URL: \url{https://en.wikipedia.org/wiki/List_of_datasets_for_machine-learning_research}

    \item \emph{Common Crawl} (a 501(c)(3) non-profit founded in 2007) maintains a free, open repository of web crawl data. They make wholesale extraction, transformation, and analysis of open web data accessible to researchers. The resulting corpus contains petabytes of data regularly collected since 2008. You may use Amazon's cloud platform to run analysis jobs directly against this data, or you can download it, whole or in part. This corpus was used to train the large language model ChatGPT.

    URL: \url{https://commoncrawl.org/}

    \item The \emph{Webis-Dataset-Reviews-21}~\cite{Kolyada2021} corpus comprises the curated list of 13,372 NLP-related datasets and their 539,411 mentions extracted from all publications available in ACL Anthology corpus.

    URL: \url{https://webis.de/data/webis-dataset-reviews-21.html}

    \item The \emph{Rapid API Hub} enables the data scientist to discover and connect to thousands of APIs within sports, finance, data, entertainment, travel, location, science, food, transportation, music, business, visual recognition, tools, text analysis, weather, gaming, SMS, events, health and fitness, and payments.

    URL: \url{https://rapidapi.com/hub}
    
\end{enumerate}

\section{Datasets}

In this section we will list some of the datasets that we (or some of our collaborators) find interesting, classifying them by area of application. Our goal is not to repeat the list of standard benchmark datasets (which you can find in Wikipedia, see ``Wikipedia's List of datasets for machine-learning research'' above). Our goal is to list those datasets that can lead to application-specific insights and, hopefully, academic and industrial breakthroughs.

\subsection{General}

\begin{enumerate}

\item \emph{U.S. Government's open data (DATA.GOV)} contains around 236,476 datasets in different fields such as agriculture, climate, education, finance, health, etc. It also has a search box that helps you to find out the data you are looking for. The datasets are public in nature. Users can download datasets in different formats. The data is maintained by the GitHub repository. Data.giv is a dataset aggregator and a home of U.S. Government's open data.

URL: \url{https://data.gov/}

\item \emph{The United Kingdom Find Open Data} --- find data published by central government, local authorities, and public bodies to help you build products and services. The search function covers business and economy, crime and justice, defence, education, environment, government, government spending, health, mapping, society, towns and cities, transport, digital service performance, government reference data.

URL: \url{https://www.data.gov.uk/}

\item \emph{The United Kingdom statistical data sets} include 1,037 data sets such as unclaimed estates list, fishing quota allocations for England and the UK, quota use statistics, quarterly traffic estimates~(TRA25), live tables on planning application statistics, historical and discontinued planning live tables, Marine Management Organisation effort statistics, non-association independent schools inspections and outcomes: management information, port and domestic waterborne freight statistics: data tables (PORT), and more.

URL: \url{https://www.gov.uk/government/statistical-data-sets}

\item \emph{The UK Data Service} is the UK's largest collection of economic, population, and social research data for teaching, learning, and public benefit. The website offers a selection of links to open data platforms, portals, and hubs. The list includes European sources, non-European sources, and non-government sources.

URL: \url{https://ukdataservice.ac.uk/}

\item \emph{Open Government Data~(OGD)} Platform India is a single point of access to datasets in open formats published by Ministries and Departments. The source consists of datasets on real-life of all shapes and sizes along with their APIs and visualisations. The datasets are available for public use.

URL: \url{https://data.gov.in/}

\item \emph{Wharton Research Data Services~(WRDS)} provide access to data from numerous data vendors. WRDS data is compiled from independent sources that specialise in specific historical data. Some sources include Capital IQ, NYSE, CRSP, and Refinitiv (formerly Thomson Reuters), and more specialised sources such as Markit, FactSet, Hedge Fund Research, Inc., Eventus, and GSIOnline with more added regularly to meet the needs of the clients. The datasets include financial statements, audit and regulatory filings, banks, segments/industry data, compensation, intellectual property, stock prices, analyst estimates, intraday trades and quotes, indices and factors, bonds and fixed income, private equity/venture capital, mutual fund / hedge fund / ETF returns, derivatives / options, REITs, currency exchange rates, ownership, mergers and acquisitions, ESG: Environmental, Social, Governance data, economics, marketing, news (including RavenPack News Analytics and SnP Capital IQ Key Developments), and healthcare.

URL: \url{https://wrds-www.wharton.upenn.edu/}

\item Microsoft along with the external research community launched a repository in July 2018 known as \emph{Microsoft Research Open Data}. It consists of curated datasets that were used in the published research studies. In addition, datasets are present in different fields such as computer science, biology, healthcare, mathematics, etc. The repository offers a wide variety of formats for downloading datasets.

URL: \url{https://www.microsoft.com/en-us/research/project/microsoft-research-open-data/}

\item \emph{Socrata OpenData} is a portal that contains multiple datasets. This broad range of information makes it more attractive and useful among data scientists and other researchers. You can look for the data in the tabular form in the browser or can use some built-in visualisation tools.

URL: \url{https://dev.socrata.com/data/}

\item \emph{Kaggle} offers more than 250,233 datasets across different subjects, many of them open. At the time of writing, the ten most popular datasets on Kaggle\footnote{\url{https://www.kaggle.com/discussions/general/260690}} are:

\begin{itemize}

\item \emph{Credit Card Fraud Detection}~\cite{Pozzolo2015} --- This dataset helps companies and teams recognise fraudulent credit card transactions. The dataset contains transactions made by European credit cardholders in September 2013. The dataset presents details of 284,807 transactions, including 492 frauds, that happened over two days.

URL: \url{https://www.kaggle.com/mlg-ulb/creditcardfraud}

\item \emph{European Soccer Database} --- The dataset contains 25,000+ matches, 10,000+ players, 11 European countries with their lead championship, seasons 2008 to 2016, players and teams' attributes sourced from EA Sports' FIFA video game series, including weekly updates, team line up with squad formation (X, Y coordinates), betting odds from up to 10 providers, detailed match events (goal types, corner, possession, fouls, etc.) for 10,000+ matches. For non-commercial use only.

URL: \url{https://www.kaggle.com/hugomathien/soccer}

\item \emph{Avocado Prices} --- The dataset shows the historical data on avocado prices and sales volume in multiple US markets. THe information has been generated from the Hass Avocado Board website. It represents weekly 2018 retail scan data for national retail volume (units and price, along with region, types (conventional or organic), and avocado sold volume. The dataset can be applied to other fruits and vegetables across geographies. Contributed by Hass Avocado Board.

URL: \url{https://www.kaggle.com/neuromusic/avocado-prices}

\item \emph{Open Food Facts} --- This is a free, open, collaborative database of food products worldwide, with ingredients, allergens, nutrition facts, and all the tidbits of information found on product labels. The database is a part of Google's Summer of Code 2018. 5,000+ contributors have added 600K+ products from 150 countries using an app or their camera to scan barcodes and upload pictures of products and their labels.

URL: \url{https://www.kaggle.com/openfoodfacts/world-food-facts}

\item \emph{IBM HR Analytics Employee Attrition and Performance} --- Created by IBM data scientists, this fictional dataset is used to predict attrition in an organisation. It uncovers various factors that lead to employee attrition and explores correlations such as ``a breakdown of distance from home by job role and attrition,'' or ``comparison of average monthly income by education and attrition.''

URL: \url{https://www.kaggle.com/pavansubhasht/ibm-hr-analytics-attrition-dataset}

\item \emph{Red Wine Quality}~\cite{Cortez2009ModelingWP} --- Red wine quality is a clean and straightforward practice dataset for regression or classification modelling. The two datasets available are related to red and white variants of the Portuguese `Vinho Verde' wine. The information in this dataset includes fixed acidity, volatile acidity, citric acid, residual sugar, chlorides, free sulfur dioxide, total sulfur dioxide, density, pH, and others. The dataset is also available on the UCI machine learning repository.

URL: \url{https://www.kaggle.com/uciml/red-wine-quality-cortez-et-al-2009}

\item \emph{Medical Cost Personal Datasets}~\cite{Lantz2019} --- This dataset is used for forecasting insurance via regression modelling. THe dataset includes age, sex, body mass index, children (dependents), smoker, region and charges (individual medical costs billed by health insurance). The dataset is also available on GitHub.

URL: \url{https://www.kaggle.com/mirichoi0218/insurance}

\item \emph{Open Food Facts} --- This is a free, open, collaborative database of food products worldwide, with ingredients, allergens, nutrition facts, and all the tidbits of information found on product labels. The database is a part of Google's Summer of Code 2018. 5,000+ contributors have added 600K+ products from 150 countries using an app or their camera to scan barcodes and upload pictures of products and their labels.

URL: \url{https://www.kaggle.com/openfoodfacts/world-food-facts}

\item \emph{Machine Learning and Data Science Survey} --- Kaggle conducted an industry-wide survey in 2017 to establish a comprehensive overview of the data science and machine learning landscape. The survey received over 16K responses gathering information around data science, machine learning innovation, how to become data scientists, and more. You can find the kernels used in the report here.

URL: \url{https://www.kaggle.com/kaggle/kaggle-survey-2017}

\item \emph{Titanic} --- The Titanic dataset consists of original data from the Titanic competition and is ideal for binary logistic regression. The dataset contains information about the passenger's id, age, sex, fare, etc. The Titanic competition involves users creating a machine learning model that predicts which passengers survived the Titanic shipwreck.

URL: \url{https://www.kaggle.com/heptapod/titanic}

\item \emph{Annotated Corpus for Named Entity Recognition} --- This dataset is extracted from the Groningen Meaning Bank~(GMB) corpus, tagged, annotated, and built specifically to train the classifier to predict labelled entities such as name, location, etc. It gives you a broad view of feature engineering and helps solve business problems like picking entities from electronic medical records, etc.

URL: \url{https://www.kaggle.com/abhinavwalia95/entity-annotated-corpus}

\end{itemize}

Further datasets can be found at the following URL: \url{https://www.kaggle.com/datasets}

\item \emph{UC Irvine Machine Learning Repository} maintains 644 datasets as a service to the machine learning community. Here you can donate and find datasets used by millions of people all around the world.

URL: \url{https://archive.ics.uci.edu/}

\item \emph{Academic Torrents} is not a mainstream yet powerful repository to share data. The main purpose behind its creation is an attempt to make academic datasets and research papers available via BitTorrent. However, the main focus is to share datasets from different research papers.

URL: \url{https://academictorrents.com/}

\item \emph{Reddit} is a popular social news site, but it also acts as a discussion board to share datasets. Such discussion boards are called \emph{subreddits} or \emph{r/datasets}. It is a place to share, find and discuss datasets. However, the quality of the datasets may vary because different users submit them.

URL: \url{https://www.reddit.com/r/datasets/}

\item \emph{Awesome Public Datasets} is a repository on GitHub of high quality topic-centric public data sources. They are collected and tidied from blogs, answers, and user responses. Almost all of these are free.

URL: \url{https://github.com/awesomedata/awesome-public-datasets}

\item \emph{Data is Plural} is a weekly newsletter of useful/curious datasets.

URL: \url{https://www.data-is-plural.com/}

\item \emph{Data World} is an open data repository containing data contributed by thosands of users and organisations all around the world. It contains hard to find data. For example, it contains 3,667 free health datasets.

URL: \url{https://data.world/}

\item \emph{Library of Congress} offers datasets as potential sources for data science or machine learning projects. Time series are available for most economic, business, census, and demographic statistics. For additional sources of datasets, see the Business Reference Services guide on Data sets (\emph{BeOnline}).

The Library of Congress makes these two datasets freely available to researcher and analysts: (1) By the People Data Sets --- transaction data was created from completed By the People campaigns and is available in bulk as zipped .csv files; (2) Web Archive Datasets --- The Library of Congress Web Archives provides users with derivative datasets for users to download, re-use, and explore.

URL: \url{https://guides.loc.gov/datasets/repositories}

\item \emph{Datarade.ai} offer an interface for finding, comparing and accessing data products from 500+ premium data providers across the globe.

URL: \url{https://datarade.ai/}

\end{enumerate}

\subsection{Finance and economics}

\begin{enumerate}

\item \emph{Bloomberg Terminal}, which the developers descibe as ``a global icon of progress'' has ``revolutionized an industry by bringing transparency to financial markets. More than four decades on, it remains at the cutting edge of innovation and information delivery --- with fast access to news, data, unique insight and trading tools helping leading decision makers turn knowledge into action.'' The Terminal provides coverage of markets, industries, companies, and securities across all asset classes. Bloomberg Terminal is known for its ``unparalleled coverage.'' Access to the Bloomberg Terminal is available on a commercial basis.

URL: \url{https://www.bloomberg.com/professional/solution/bloomberg-terminal/}

\item \emph{Bloomberg Professional Services} offer several APIs for accessing Bloomberg data. For more information refer to the BLPAPI Developer's Guide --- a tutorial for developing applications with BLPAPI in C++, Java, and .NET. There is API Windows, API Linux, API macOS, Schema Downloader, and API Python.

URL: \url{https://www.bloomberg.com/professional/support/api-library/}

\item \emph{Bloomberg Server API (SAPI)} delivers a powerful complement to the Bloomberg Terminal. It allows the user to consume Bloomberg's real-time market, historical, and key reference data, as well as calculation engine capabilities when using proprietary and third-party applications.

URL: \url{https://www.bloomberg.com/professional/product/server-api/}

\item \emph{Refinitiv Eikon} is the financial analysis desktop and mobile solution for access to leading data and content, Reuters news, market data, and liquidity pools. Access to Refinitiv Eikon is available on commercial basis.

URL: \url{https://www.refinitiv.com/en/products/eikon-trading-software}

\item Refinitiv offers numerous \emph{Refinitif APIs} for accessing data programmatically, such as App Studio --- Web SDK, CIAM, Cash RFQ FIX API, and more.

URL: \url{https://developers.refinitiv.com/en/api-catalog}

\item The curators claim that ``The world's most powerful data lives on \emph{Quandl}.'' It is the premier source for financial, economic, and alternative datasets, serving investment professionals. Quandl's platform is used by over 400,000 people, including analysts from the world's top hedge funds, asset managers, and investment banks.

Quandl is subdivided into Core Financial Data --- market data from hundreds of sources via API, or directly into Python, R, Excel and many other tools; and Alternative Data for institutional clients only: ``We bring undiscovered data from non-traditional publishers to investors seeking unique, predictive insights.''

URL: \url{https://demo.quandl.com/}

\item \emph{Wharton Research Data Services~(WRDS)} provide access to data from numerous data vendors. WRDS data is compiled from independent sources that specialise in specific historical data. Some sources include Capital IQ, NYSE, CRSP, and Refinitiv (formerly Thomson Reuters), and more specialised sources such as Markit, FactSet, Hedge Fund Research, Inc., Eventus, and GSIOnline with more added regularly to meet the needs of the clients. The datasets include financial statements, audit and regulatory filings, banks, segments/industry data, compensation, intellectual property, stock prices, analyst estimates, intraday trades and quotes, indices and factors, bonds and fixed income, private equity/venture capital, mutual fund / hedge fund / ETF returns, derivatives / options, REITs, currency exchange rates, ownership, mergers and acquisitions, ESG: Environmental, Social, Governance data, economics, marketing, news (including RavenPack News Analytics and SnP Capital IQ Key Developments), and healthcare.

URL: \url{https://wrds-www.wharton.upenn.edu/}

\item \emph{Trading Economics} provides its members with access to millions of economics indicators for 196 countries and historical/delayed/live quotes for exchange rates, stocks, indexes, bonds, and commodity prices.

URL: \url{https://tradingeconomics.com/analytics/features.aspx?source=footer}

\item \emph{BMLL Technologies} is a leading, independent provider of Level 3 historical data and analytics. The company claims that their Level 3 data is ``the cleanest order book data available anywhere in the capital markets.'' BMLL Level 3 data captures every order sent to the market, and is fully harmonised across venues and asset classes. BMLL's 6+ years of data ana analytics span global equities, ETFs and futures from 75+ trading venues, and are used by banks and brokers, asset managers, global exchange groups, and hedge funds.

URL: \url{https://www.bmlltech.com/}

\item \emph{Databento} provides real-time and historical market and reference data, sourced directly from colocation sites. At present the database covers equities, equity options, futures, and options on futures. The APIs provided by Databento (WebSocket, Raw, HTTP, C++, Python) offer both historical and live data. The data covers a large

URL: \url{https://databento.com/}

\item \emph{FirstRate Data} is a leading provider of high-resolution intraday stock market, crypto, futures, and FX data. They source their historical stock data directly from major exchanges and fully adjust the data for both splits and dividends. Futures and ETF datasets are also sourced from co-located servers in major exchanges. All datasets are rigorously tested for accuracy. The historical intraday data solutions are research-ready and are offered in 1-minute, 5-minute, 30-minute, 1-hour, and 1-day intraday stock data as well as intraday futures, ETFs, and FX data going back 15 years, and tick data going back 10 years.

URL: \url{https://firstratedata.com/}

\item \emph{Turnleaf Analytics} use machine learning, machine learning, alternative data and new technologies to create economic forecasts of inflation and analyze financial markets, in order to provide our clients the much needed clarity required when navigating the complex financial landscape.

URL: \url{https://turnleafanalytics.com/}

\item \emph{FINRA} provides real-time and historic data for most \emph{TRACE-eligible securities} (including US corporate bonds) to members and any others that choose to subscribe for a fee. The data feeds include real-time data, end-of-day data, terminals and snapshot data.

URL: \url{https://www.finra.org/filing-reporting/trace/data}

\item \emph{Neptune} deliver targeted, high quality data directly from corporate bond dealers into core workflow tools. Real-time, structured and standardised connectivity means reduction of ``noise'' and inaccurate data. The use of FIX allows for ease of connectivity via API, OMS/EMS or via Neptune's GUI. High quality, pre-trade bond data is available via one-connection, from the leading sell-side market makers in fixed income.

URL: \url{https://neptunefi.com/}

\item \emph{Bonds.com} provide market and static data for over 20,000 bonds. The data consists of real-time tick-by-tick (top of book or full depth, 30-100mm+ updates per day), intra-day intervals, end of day, and historical.

URL: \url{https://bonds.com/data/}

\item \emph{Dukascopy Swiss Banking Group} provide the \emph{Historical Data Feed}, which includes historical price data for a variety of financial instruments (e.g. FX, commodities, and indices):

\url{https://www.dukascopy.com/swiss/english/marketwatch/historical/}

\item \emph{Investing.com} offer live FX option volatility surfaces for G10 and EM currency pairs.

URL: \url{https://www.investing.com/currencies/forex-options}

\item \emph{Google Finance} provides free real-time quotes, international exchanges, up-to-date financial news and analytics.

URL: \url{https://www.google.com/finance/?hl=en}

\item \emph{Yahoo Finance} provides free stock quotes, up-to-date news, portfolio management resources, international market data, social interaction, and mortgage data.

URL: \url{https://finance.yahoo.com/}

\item \emph{Credit Card Fraud Detection}~\cite{Pozzolo2015} --- This dataset helps companies and teams recognise fraudulent credit card transactions. The dataset contains transactions made by European credit cardholders in September 2013. The dataset presents details of 284,807 transactions, including 492 frauds, that happened over two days.

URL: \url{https://www.kaggle.com/mlg-ulb/creditcardfraud}

\end{enumerate}

\subsection{Legal (laws and regulations)}

\begin{enumerate}
\item \emph{The United States Patent and Trademark Office (UPSTO)} Patent Public Search tool is a web-based patent search application that has replaced the internal legacy search tools PubEast and PubWest and external legacy search tools PatFT and AppFT.

URL: \url{https://www.uspto.gov/patents/search}

\item \emph{The United States Patent and Trademark Office (UPSTO)} new trademark search system will soon replace the existing Trademark Electronic Search System (TESS):

URL: \url{https://www.uspto.gov/trademarks/search}

\item \emph{GOV.UK Search-for-a-patent} searches for published patent applications and registered patents using the Intellectual Property Office's patent information and document service (Ipsum) and patent publication service.

URL: \url{https://www.gov.uk/search-for-patent}

\item \emph{GOV.UK Search-for-a-trade-mark} can be used to search for a UK trae mark by trade mark number, owner, keyword, phrase, or image.

URL: \url{https://www.gov.uk/search-for-trademark}

\item For trade marks in Jersey, search the \emph{Jersey trade mark register}.

URL: \url{http://www.jgreffe-online.gov.je/trademarksdb/searchform.asp}

\item For trade marks in Guernsey, search the \emph{Guernsey trade mark register}.

URL: \url{http://ipo.guernseyregistry.com/article/107508/View-the-Trade-Mark-Register}

\item \emph{Espacenet} provides free access to over 140 million patent documents of the European Patent Office.

URL: \url{https://worldwide.espacenet.com/}

\item \emph{Deutsches Patent- und Merkenamt (DPMA)} offers access to patents, trademarks, and designs via DMPAregister, DEPATISnet, DPMAdirektWeb, and DPMAdirektPro.

URL: \url{https://www.dpma.de/}

\item The \emph{China National Intellectual Property Administration~(CNIPA)} offers patent and trademark search.

URL: \url{https://english.cnipa.gov.cn/}

\item The \emph{Indian Patent Advanced Search System of Intellectual Property India} offers patent search.

URL: \url{https://iprsearch.ipindia.gov.in/publicsearch}

\item Using \emph{World Intellectual Property Organization (WIPO) PATENTSCOPE} you can search 113 million patent documents including 4.7 million published international patent applications (PCT).

URL: \url{https://patentscope.wipo.int/search/en/search.jsf}

\item \emph{Google Patents} searches and displays the full text of patents from around the world.

URL: \url{https://patents.google.com/}
\end{enumerate}

\subsection{Life sciences}

\begin{enumerate}

\item The World Health Organization~(WHO) \emph{International Clinical Trials Registry Platform~(ICTRP)} is ``a voluntary platform to link clinical trials registers in order to ensure a single point of access and the unambiguous identification of trials with a view to enhancing access to information by patients, families, patient groups and others.'' The platform has a searchable portal: International Clinical Trials Registry Platform~(ICTRP) search portal.

URL: \url{http://apps.who.int/trialsearch/}

\item \emph{ClinicalTrials.gov} is a registry and results database of privately and publicly funded clinical studies conducted around the world. The resource is provided by the U.S. National Library of Medicine. Each study record includes a summary of the study protocol. Some study records include a summary of the results in a tabular format. Studies can be searched by status, condition or disease, contry, or by other terms. The database is continually updated; it currently lists over 300,000 research studies located in all 50 states in the United States of America and over 200 other countries adound the world.

URL: \url{https://clinicaltrials.gov/}

\item The \emph{ISRCTN registry} is a primary clinical trial registry. It is recognised by the World Health Organisation~(WHO) and the International Committee of Medical Journal Editors~(ICMJE) and accepts all clinical research studies (whether proposed, ongoing or completed). Each study record includes a plain English summary as well as details of the study protocol. All study records can be searched using an advanced search function. Currently the registry lists over 18,000 studies.

URL: \url{https://www.isrctn.com/}

\item The \emph{EU Clinical Trials Register} contains information on interventional clinical trials on medicines conducted in the European Union~(EU), or the European Economic Area~(EEA) which started after 1 May 2004. It also includes some clinical trials conducted outside the EU/EEA or some older trials that meet certain criteria. Study records indicate the trial protocol and provide results where available. Records can be searched using an advanced search function. Currently the registry displays over 34,000 clinical trials.

URL: \url{https://euclinicaltrials.eu/}

\item The \emph{Pan African Clinical Trail Registry~(PACTR)} is a regional register of clinical trials conducted in Africa. An open-access platform where clinical trials can be registered free of charge, providing an electronic database of planned trials and trials currently in progress.

URL: \url{https://pactr.samrc.ac.za/}

\item \emph{PubChem}~\cite{Kim2022} is the world's largest collection of freely accessible chemical information. Users can search chemicals by name, molecular formula, structure, and other identifiers. The database contains chemical and physical properties, biological activities, safety and toxicity information, patents, literature citations, and more.

PubChem contains 116 million compounds, 308 million substances, and 934 data sources.

URL: \url{https://pubchem.ncbi.nlm.nih.gov/}

\item \emph{ChEMBL} or \emph{ChEMBLdb}~\cite{Mendez2018} is a manually curated chemical database of bioactive molecules with drug inducing properties. It is maintained by the European Bioinformatics Institute, of the European Molecular Biology Laboratory, based at the Welcome Trust Genome Campus, Hinxton, UK. The database brings together chemical, biological, and genomic data to aid the translation of genomic information into effective new drugs.

At the time of writing, ChEMBL includes 15,398 targets, 2,399,743 distinct compounds, 20,334,684 activities, 88,630 publications, and 215 deposited datasets.

URL: \url{https://www.ebi.ac.uk/chembl/}

\item \emph{Chemical Entities of Biological Interest}~\cite{Hastings2015}, also known as \emph{ChEBI}, is a chemical database and ontology of molecular entities focused on `small' chemical compounds that is part of the Open Biomedical Ontologies effort at the European Bioinformatics Institute~(EBI).

In order to create ChEBI, data from a number of sources were incorporated and subjected to merging procedures to eliminate redundancy.

Four of the main sources from which the data are drawn are:

\begin{itemize}
    \item \emph{IntEnz} --- the Integrated relational Enzyme database of the EBI. IntENz is the master copy of the Enzyme Nomenclature, the recommendations of the NC-IUBMB on the Nomenclature and Classification of Enzyme Catalysed Reactions.
    \item \emph{KEGG COMPOUND} --- One part of the Kyoto Encyclopedia of Genes and Genomes LIGAND database, COMPOUND is a collection of biochemical compound structures.
    \item \emph{PDBeChem} --- The service providing web access to the Chemical Component Dictionary of the wwPDB as this is loaded into the PDBe database at the EBI.
    \item \emph{ChEMBL} --- A database of bioactive compounds, their quantitative properties and bioactivities, abstracted from the primary scientific literature. It is part of the ChEMBL resources at the EBI.
\end{itemize}

URL: \url{https://www.ebi.ac.uk/chebi/}

\item \emph{DrugBank}~\cite{Wishart2017} is the most comprehensive, up-to-date, and accurate drug database on the market. It is available for commercial and academic research and is also accessible via a clinical API.

DrugBank offers customisable drug search options, drug-drug interaction checker, allergy and cross-sensitivities information, and US drug labels.

DrugBank Online is offered to the public as a free-to-access resource. Use and re-distribution of the content of DrugBank Online or the DrugBank Data, in whole or in part, for any purpose requires a license. Academic users may apply for a free license for certain use cases and all other users require a paid license.

DrugBank boasts 26,500+ citations in scientific publications, 13 of 20 top pharma companies are customers, and more than 1.5 billion USD has been spent on research utilising DrugBank.

URL: \url{https://www.drugbank.com/}

\item \emph{MoleculeNet}~\cite{Wu2017} is a benchmark specially designed for testing machine learning methods of molecular properties. As we aim to facilitate the development of molecular machine learning method, this work curates a number of dataset collections, creates a suite of software that implements many known featurizations and previously proposed algorithms. All methods and datasets are integrated as parts of the open source DeepChem package (MIT license).

MoleculeNet is built upon multiple public databases. The full collection currently includes over 700,000 compounds tested on a range of different properties. We test the performances of various machine learning models with different featurizations on the datasets(detailed descriptions here), with all results reported in AUC-ROC, AUC-PRC, RMSE and MAE scores.

URL: \url{https://moleculenet.org/}

\item \emph{AlphaFold Protein Structure Database}~\cite{Jumper2021} provides open access to over 200 million protein structure predictions to accelerate scientific research. AlphaFold is an AI system developed by DeepMind that predicts a protein's 3D structure from its amino acid sequence. It regularly achieves accuracy competitive with experiment. DeepMind and EMBL's European Bioinformatics Institute (EMBL-EBI) have partnered to create AlphaFold DB to make these predictions freely available to the scientific community. The latest database provides broad coverage of UniProt (the standard repository of protein sequences and annotations). They provide individual downloads for the human proteome and for the proteomes of 47 other key organisms important in research and global health. They also provide a download for the manually curated subset of UniProt (Swiss-Prot).

URL: \url{https://alphafold.ebi.ac.uk/}

\item \emph{RCSB Protein Data Bank (RCSB PDB)}~\cite{Berman2000} enables breakthroughs in science and education by providing access and tools for exploration visualisation, and analysis of (1) experimentally-determined 3D structures from the Protein Data Bank~(PDB) archive; (2) Computed Structure Models~(CSM) from AlphaFold DB and Model Archive. These data can be explored in context of external annotations providing a structural view of biology.

URL: \url{https://www.rcsb.org/}

\item \emph{KEGG COMPOUND}~\cite{Hashimoto2008} is one of the four original databases, together with KEGG PATHWAY, KEGG GENES and KEGG ENZYME, introduced at the start of the KEGG project in 1995. It is a collection of small molecules, biopolymers, and other chemical substances that are relevant to biological systems. Each entry is identified by the C number, such as C00047 for L-lysine, and contains chemical structure and associated information, as well as various links to other KEGG databases and outside databases. Some COMPOUND entries are also represented as GLYCAN and DRUG entries with the ``Same as'' links.

While GLYCAN entries are represented as tree structures with monosaccharide codes, COMPOUND entries for peptides and polyketides, such as C11996 for methymycin, are represented as sequences using the abbreviation codes for the monomeric units of amino acids and carboxylic acids.

URL: \url{https://www.genome.jp/kegg/compound/}

\item \emph{ZINC20}~\cite{Irwin2012} is a free database of commercially-available compounds for virtual screening. ZINC contains over 230 million purchasable compounds in ready-to-dock, 3D formats. ZINC also contains over 750 million purchasable compounds you can search for analogs in under a minute.

ZINC is provided by the Irwin and Shoichet Laboratories in the Department of Pharmaceutical Chemistry at the University of California, San Francisco (UCSF). We thank NIGMS for financial support (GM71896).

URL: \url{https://zinc20.docking.org/}

\item The \emph{Tox21}~\cite{Richard2020} dataset comprises 12,060 training samples and 647 test samples that represent chemical compounds. There are 801 ``dense features'' that represent chemical descriptors, such as molecular weight, solubility, or surface area, and 272,776 ``sparse features'' that represent chemical substructures (ECFP10, DFS6, DFS8; stored in Matrix Market Format). Machine learning methods can either use sparse or dense data or combine them. For each sample there are 12 binary labels that represent the outcome (active/inactive) of 12 different toxicological experiments. Note that the label matrix contains many missing values (NAs).

URL: \url{https://tripod.nih.gov/tox21/challenge/}

\item \emph{FooDB} is the world's largest and most comprehensive resource on food constituents, chemistry, and biology. It provides information on both macrunutrients and micronutrients, including many of the constituents that give foods their flavour, colour, taste, texture, and aroma. Each chemical entry in the FooDB contains more than 100 separate data fields covering detailed compositional, biochemical and physiological information (obtained from the literature). This includes data on the compound's nomenclature, its description, information on its structure, chemical class, its physico-chemical data, its food source(s), its colour, its aroma, its taste, its physiological effect, presumptive health effects (from published studies), and concentrations in various foods. Users are able to browse or search FooDB by food source, name, descriptors, function or concentrations. Depending on individual preferences users are able to view the content of FooDB from the Food Browse (listing foods by their chemical composition) or the Compound Browse (listing chemicals by their food sources).

URL: \url{https://www.foodb.ca/}

\item \emph{Open Food Facts} --- This is a free, open, collaborative database of food products worldwide, with ingredients, allergens, nutrition facts, and all the tidbits of information found on product labels. The database is a part of Google's Summer of Code 2018. 5,000+ contributors have added 600K+ products from 150 countries using an app or their camera to scan barcodes and upload pictures of products and their labels.

URL: \url{https://www.kaggle.com/openfoodfacts/world-food-facts}

\item \emph{Scopus} is Elsevier's abstract and citation database launched in 2004. Scopus is curated by independent subject matter experts. As of March 2023, Scopus includes 27,950 active titles: 26,591 active peer-reviewed journals, 192 trade journals, 1,167 book series, 11.7+ million conference papers from 148,500+ worldwide events, ``articles-in-press'' from 9,100+ journals, 292,000+ stand-alone books, 90.6+ million records (84+ million records post-1969 with references and 6.5+ million records pre-1970 with the oldest record dating back to 1788). There are also 49.2+ million patent records from five patent offices.

URL: \url{https://www.scopus.com/}

\item \emph{Medical Cost Personal Datasets}~\cite{Lantz2019} --- This dataset is used for forecasting insurance via regression modelling. THe dataset includes age, sex, body mass index, children (dependents), smoker, region and charges (individual medical costs billed by health insurance). The dataset is also available on GitHub.

URL: \url{https://www.kaggle.com/mirichoi0218/insurance}

\item \emph{Plants For A Future~(PFAF)} is a database of 8,000+ edible and medicinal plants. It includes hardiness zones, care, hazards, physical characteristics, synonyms, habitats, edible uses, medicinal uses, other uses, cultivation details, propagation, and more.

URL: \url{https://pfaf.org/user/}

\item The \emph{Medicinal Plant Database of the Botanical Survey of India} covers plants that are employed in different medicinal systems and ethnic medicines. India has a rich tradition of herbal medicines and it has made contributions not only in the form of Ayurveda and Siddha but also in the discovery of modern drugs and pharmacological research. This database provides information on scientific name, family, vernacular name, medicinal uses, location of species and images of herbarium specimen. In the first phase, a total of 1,915 species are listed and about 1,000 will be added in the next phase.

URL: \url{https://bsi.gov.in/page/en/medicinal-plant-database}

\item \emph{CAB Database of Plant Science} contains abstracts of internationally published scientific research.

URL: \url{http://www.cabi.org/}

\item \emph{Dr Duke's Phytochemical and Ethnobotanical Databases} is a database of the ethnobotanical uses and chemical activities in plants.

URL: \url{https://phytochem.nal.usda.gov/phytochem/search}

\item \emph{Food and Agriculture Organization of the United Nations~(FAO)} is a specialised agency of the United Nations that leads international efforts to defeat hunger. The goal is to achieve food security for all and ensure that people have regular access to enough high quality food to lead active, healthy lives.

URL: \url{http://www.fao.org/home/en/}

\item \emph{Harvard University Herbaria and Libraries}.

URL: \url{http://huh.harvard.edu/}

\item The American Botanical Council's \emph{HerbMed} and \emph{HerbMedPro} --- an interactive, electronic herbal database that provides hyperlinked access to the scientific data underlying the use of herbs for health. It is an evidence-based information resource for professionals, researchers, and general public. HerbMedPro is the professional version of HerbMed. This enhanced version provides access to the entire database with continuous updating as the information is being compiled.

URL: \url{https://www.herbalgram.org/resources/herbmedpro/}

\item \emph{Plantes m{\'e}dicinales}, the journal of \emph{Guilde des herboristes}.

URL: \url{http://www.guildedesherboristes.org/}

\item The \emph{Integrative Medicine Program} at the MD Anderson Cancer Center engages patients and their families to become active participants in improving their physical, psycho-spiritual, and social health. The ultimate goals are to optimise health, quality of life, and clinical outcomes through personalised evidence-based clinical care, exceptional research and education.

URL: \url{http://www.mdanderson.org/education-and-research/departments-programs-and-labs/programs-centers-institutes/integrative-medicine-program/index.html}

\item \emph{NAPRALERT} --- A relational database of all natural products, including ethnomedical information, pharmacological and biochemical information of extracts of organisms in virto, in situ, in vivo, in humans (case reports, non-clinical trials) and clinical studies. Similar information is available for secondary metabolites from natural sources.

URL: \url{http://www.napralert.org/}

\item The US National Library of Medicine (NLM), on the campus of the National Institutes of Health in Bethesda, Maryland, has been a centre of information innovation since its founding in 1836. The world's largest biomedical library, NLM maintains and makes available a vast print collection and produces electronic information resources on a wide range of topics that are searched billions of times each year by millions of people around the globe. It also supports and conducts reseearch, development, and training in medical informatics and health information technology.

URL: \url{https://www.nlm.nih.gov/}

\item The W3 Tropicos database links over 1.38M scientific names with over 6.85M specimens and over 1.55M digital images. The data includes over 165K references from over 54.9K publications offered as a free service to the world's scientific community.

URL: \url{https://www.tropicos.org/}

\item The \emph{American Botanical Council} maintains a list of databases and data sources.

URL: \url{https://www.herbalgram.org/resources/related-links-page/databases/}

\item \emph{Abbott FreeStyle Libre 3} is the world's smallest, thinnest glucose sensor that noninvasively tracks glucose levels in the body. It provides unsurpased 14-day accuracy and optimal glucose alarms, but also evolves the portfolio with new features, such as continuous real-time glucose readings automatically delivered to a person's smartphone every minute and a sensor that is easy to apply with a one-piece applicator.

URL: \url{https://www.abbott.com/corpnewsroom/strategy-and-strength/freeStyle-libre-3-worlds-smallest-sensor-is-here.html}

\item \emph{mymonX} combines a smart wearable and an app to produce non-invasive medical grade measurements of blood glucose levels, heart rate, ECG, blood pressure, oxygenation (SPo2), breathing rate (Rr), sleep, activity, and more.

URL: \url{https://mymonx.co/products/mymonx-original-smart-watch}

\item \emph{ZOE} consists of easily applicable at-home tests that give insight into blood fat, blood sugar, and gut microbiome health. The results are then mapped into ZOE Scores for food (from 0 to 100). The product is backed by several academic publications.

URL: \url{https://zoe.com/how-it-works}

\end{enumerate}

\subsection{News sentiment and social media}

\begin{enumerate}

\item \emph{RavenPack} offer news analytics, regulatory filings, earnings dates, training data, job analytics, transcripts, and insider transactions datasets.

URL: \url{https://www.ravenpack.com/}

\item \emph{Bloomberg Professional Services} supply news and social sentiment data.

URL: \url{https://www.bloomberg.com/professional/sentiment-analysis-white-papers/}

\item \emph{Reuters/Refinitiv News Sentiment} can be computed with Eikon Data APIs.

URL: \url{https://developers.refinitiv.com/en/article-catalog/article/introduction-news-sentiment-analysis-eikon-data-apis-python-example}

\item \emph{InfoTrie FinSentS} stands for Financial News and Sentiment Screener. Get real--time analysis information for over 100,000 stocks, topics, companies, people, and other assets with up to 15 years of history.

URL: \url{https://infotrie.com/finsents-stock-and-sentiment-screener/}

\item The \emph{Twitter API} enables programmatic access to Twitter.

URL: \url{https://developer.twitter.com/en/docs/twitter-api}

\item \emph{LinkedIn} offers several APIs.

URL: \url{https://developer.linkedin.com/product-catalog}

\item \emph{Meta (Facebook)} offer several APIs, platforms, products, and SDKs.

URL: \url{https://developers.facebook.com/docs/}

\end{enumerate}

\subsection{Retail and ecommerce}

\begin{enumerate}

\item The \emph{E-Commerce Sales Data} on Kaggle is a comprehensive dataset with sales data across channels and financial information. Data includes SKUs, design numbers, stock levels, product categories, product sizes, product colours, the amount paid, rate per piece, date of sale, gross amounts, and more.

URL: \url{https://www.kaggle.com/datasets/thedevastator/unlock-profits-with-e-commerce-sales-data}

\item The \emph{Electronic Product Pricing} dataset on Kaggle offers 10 fields of pricing information for 7,000 electronic products.

URL: \url{https://www.kaggle.com/datasets/retailrocket/ecommerce-dataset}

\item \emph{Datos} offer a structured data feed with information on click events and funnel actions for the most popular U.S. online stores to make it easy to track and analyse purchase funnel activity on any given retail site. This feed is enriched with retailer metadata, such as item price, product name, category, etc. for an easy to understand taxonomy of what is happening on any given retailer. Currently the dataset includes U.S. data for Amazon, Walmart, Target, and Etsy, URL categorisation by event type (search, product view, purchase, etc.), breakdown by country and plarform type --- desktop and mobile, and hundred of millions events per month. The dataset is available on commercial basis.

URL: \url{https://datos.live/online-retail-feed/}

\item \emph{Avocado Prices} --- The dataset shows the historical data on avocado prices and sales volume in multiple US markets. THe information has been generated from the Hass Avocado Board website. It represents weekly 2018 retail scan data for national retail volume (units and price, along with region, types (conventional or organic), and avocado sold volume. The dataset can be applied to other fruits and vegetables across geographies. Contributed by Hass Avocado Board

URL: \url{https://www.kaggle.com/neuromusic/avocado-prices}

\item \emph{Red Wine Quality}~\cite{Cortez2009ModelingWP} --- Red wine quality is a clean and straightforward practice dataset for regression or classification modelling. The two datasets available are related to red and white variants of the Portuguese `Vinho Verde' wine. The information in this dataset includes fixed acidity, volatile acidity, citric acid, residual sugar, chlorides, free sulfur dioxide, total sulfur dioxide, density, pH, and others. The dataset is also available on the UCI machine learning repository.

URL: \url{https://www.kaggle.com/uciml/red-wine-quality-cortez-et-al-2009}

\item \emph{RetailNext} offers accurate foot traffic measurement (Traffic 2.0), video security, occupancy, and shopper journey insights. Some of the products are built on RetailNext's Aurora, the next-generation sensor for physical location analytics.

URL: \url{https://retailnext.net/}

\item Rather than being a dataset vendor, \emph{Datarade} offer ``the easy way to find, compare, and access data products from 500+ premium data providers across the globe.''

URL: \url{https://datarade.ai/}

\item The \emph{Amazon Selling Partner API (SP-API)} is a REST-based API that helps Amazon selling partners programmatically access their data on orders, shipments, payments, and much more. Applications using the SP-API can increase selling efficiency, reduce labour requirements, and improve response time to custmers, helping selling partners grow their businesses.

Amazon's Selling Partner's API can be used for both Selling Partners and Vendors, and is designed to improve efficiency and aid in accelerating growth.

URL: \url{https://developer.amazonservices.com/}

\end{enumerate}

A number of interesting retail and ecommerce-related datasets can be found on Kaggle.

\subsection{Satellite imagery}

\begin{enumerate}

\item \emph{EarthScope Consortium} operates the National Science Foundation's Geodetic Facility for the Advancement of Geoscience (GAGE) and Seismological Facility for the Advancement of Geoscience (SAGE). The Synthetic Aperture Radar (SAR) data available from the GAGE Facility includes satellite-transmitted and received radar scans of the Earth's surface. SAR data, analysed using Interferometric SAR (InSAR) technqiues, can be used to model millimeter-to-centimeter scale deformation of the Earth's surface over regions tens to hundreds of kilometers across. These displacement fields are essential guides for studies of tectonics, earthquake focal mechanisms, volcano behaviour, hydrology, and public safety related to Earth hazards.

The primary tool for discovering and accessing SAR data from the GAGE Facility is the Seamless SAR Archive (SSARA). This tool is available as a command line utility (API) and web-based interface (GUI) that allows for the search and download of data from WInSAR's collection as well as data from the Alaska Satellite Facility (ASF) and some PI specific collections.

URL: \url{https://www.unavco.org/geodetic-imaging/sar-data/}

\item \emph{ArcGIS Living Atlas of the World} is the foremost collection of geographic information from around the globe. It includes maps, apps, and data layers. It includes Landsat Level-2 archive --- 41 years of scientific earth observation imagery. The Living Atlas Landsat Level-2 image service provides seamless access to a unque historical record of the Earth. ArcGIS Living Atlas of the World includes authoritative live feeds and other content that helps learn more about active hurricanes, cyclones, and typhoons. Since its initial release in 2022, the Wildfire Aware application in Living Atlas has been helping improve awareness and understanding of wildfires throughout the United States. ArcGIS Living Atlas also includes authoritative content that helps users learn more about current sea, temperature, and coral bleaching.

URL: \url{https://livingatlas.arcgis.com/en/home/}

\item \emph{Maxar} have decades of experience manufacturing communication and Earth observation satellites --- with more than 285 Maxar-built satellites and 2,750 cumulative years on orbit. Maxar's Earth observation constellation offers the most comprehensive suite of commercial satellite imagery; they offer diversity in resolution, currency, spectral bands, and accuracy.

URL: \url{https://www.maxar.com/products/}

\item The \emph{ICEYE} radar satellite constellation delivers radar imaging that makes it possible to see the surface of the Earth through clouds and even in total darkness. Governments and businesses can now look at their locations of interest 24/7. The constellation provides new images of the same location every hour. Tracking all changes that happen even within individual days is finally possible.

URL: \url{https://www.iceye.com/}





















\end{enumerate}

\subsection{Shipping and logistics}

\begin{enumerate}

\item \emph{Pole Star API} offers a customisable data feed that can be used to access vessel registration data for screening and tracking; retrieving screening results, vessel details; retrieving detailed watchlist checks on vessels, companies, and associated countries; retrieving detailed ship movement history; retrieving PurpleTRAC PDF Screening Report; monitor tracked vessels and getting their positions and other events.

URL: \url{https://developers.polestar-production.com/getting-started}

\item \emph{Spire Maritime} offer \emph{Enhanced Satellite Automatic Identification System~(AIS)}, a solution that offers vessel tracking in highly conested areas (busy shipping lanes or congested ports) where signal collision makes the vessel signal detection harder for other AIS collection methods. Such areas include the Arabian Gulf, Bab al-Mandab Strait, Cape of Good Hope, Gulf of Mexico, Mediterranean Sea, North and Baltic Seas, South China Sea, Strait of Gibraltar, and Suez Canal. Enhanced Satellite AIS provides a high frequency of position updates and global AIS coverage that helps customers track the vessels with enhanced detection. The product comes with a live API feed and historical data.

URL: \url{https://spire.com/maritime/solutions/enhanced-satellite-ais/}

\item \emph{Spire Maritime's Port Events} enable effective supply chain monitoring, port operations, and feeding data science models. Port Events is based on the most comprehensive AIS coverage available, including Satellite AIS, Terrestrial AIS, and Enhanced Satellite AIS.

URL: \url{https://spire.com/maritime/solutions/port-events/}

\item \emph{Starboard Maritime Intelligence} helps nations tackle complex maritime challenges, ranging from risk assessing arriving vessels to detecting illegal fishing and uncovering non-reporting dark vessels. By combining global Automatic Identification System~(AIS) data, multiple layers of satellite data, scientific models, and other information or intelligence, Starboard enables teams to effectively analyse and investigate vessels and areas --- all on a secure and intuitive platform.

URL: \url{https://starboard.nz/}

\item \emph{Critchlow Geospatial} offer next-generation satellite imagery (including vast imagery archives) for commercial use integrated with GIS software and authoritative location-based data.

URL: \url{https://www.critchlow.co.nz/}

\item \emph{Al{\'e}n Space} provides an end-to-end solution to orbit clients' Automatic Identification System~(AIS) services. They help companies interested in providing maritime security services, law-inforcement, Search and Rescue~(SAR), maritime surveillance, environmental solutions, and fleet management services for commercial users (shipping companies and ship owners).

SAT-AIS is a solution that overcomes terrestrial coverage limitations with the potential to provide AIS services for any given area on Earth. The VDES solution (VHF Data Exchange System) with small satellites takes advantage of the new VDE-SAT functionality, which will allow bidirectional communications and a unified standard in maritime communications.

The VDES satellite service (VDE-SAT) is prepared to transform the sector and to meet the needs of the maritime industry. VDES will allow real-time control of all data flows between vessels, with authorities, and service providers around the world.

Vessels that are not broadcasting their identification, position, and course with AIS transponders can be detected with SIGINT capabilities that rely on CubeSats. Those dark vessels are still communicating with puth-to-talk (PTT) radio systems or satellite phones, or they are navigating with S-band or X-band radio systems. Those signals can be identified from space.

URL: \url{https://alen.space/small-satellites-for-ais-services/}

\item \emph{ICEYE} also offer a SAR satellite data solution that can be used to surveil maritime activity in any area of interest --- day or night, in any weather, which can be used quickly to react to potential security threats or illegal activities. Such data can be used to detect dark vessels and oil trafficking from space.

URL: \url{https://www.iceye.com/sar-data-applications/maritime-domain-awareness}

\item \emph{Unseenlabs}, a startup commercial company of Rennes in northwestern France, provides the civilian and military in the maritime sector with the ability to locate ships by detecting and characterising their passive electromagnetic signature. They claim to be ``able to track any ship, anywhere, anytime, where other systems cannot.'' The company describes itself as the world leader in radio frequency data and solutions for maritime domain awareness.

URL: \url{https://unseenlabs.space/}

















\item \emph{SEA.AI} detects floating objects early, using thermal and optical cameras to catch even objects that escape conventional systems such as Radar or AIS: Unsignalled crafts or other floating obstacles, e.g., containers, tree trunks, buoys, inflatables, kayaks, persons over board, etc. SEA.AI offer SEA.AI Offshore (high-tech safety and convenience for blue water sailors), SEA.AI Sentry (for commercial and government use as well as for use on motoryachts), SEA.AI Competition (for ocean racing and performance yachts with rotating mast). The SEA.AI system computes input from lowlight and thermal cameras, using latest Machine Vision technology, best-in-class deep learning capabilities, and a proprietary database of millions of annotated marine objects.

URL: \url{https://sea.ai/}

\item \emph{Visivise container tracking tool} collects data from multiple sources such as shipping lines, AIS data, port and terminals, railways, and then refine and standardise them to make a unified experience to track cargos and bring visibility.

URL: \url{https://www.visiwise.co/tracking/container/}

\item \emph{MYTRACKINGDEVICES GPS and IoT platform} provides solutions for shipment tracking and monitoring, asset tracking and recovery, vehicle tracking, and personal tracking. The solution consists of small tracking devices that transmit their location from anywhere in the globe by connecting to cellular data networks. The low-powered devices are capable of up to 12 months of battery life and even longer for tracking containerised cargo. The current precision locating technology combines multiple sensors using GPS, WiFi, and Cell-ID.

URL: \url{https://mytrackingdevices.com/gps-container-tracking/}

\item \emph{Titanic} --- The Titanic dataset consists of original data from the Titanic competition and is ideal for binary logistic regression. The dataset contains information about the passenger's id, age, sex, fare, etc. The Titanic competition involves users creating a machine learning model that predicts which passengers survived the Titanic shipwreck.

URL: \url{https://www.kaggle.com/heptapod/titanic}

\end{enumerate}

\subsection{Sports}

\begin{enumerate}

\item \emph{European Soccer Database} --- The dataset contains 25,000+ matches, 10,000+ players, 11 European countries with their lead championship, seasons 2008 to 2016, players and teams' attributes sourced from EA Sports' FIFA video game series, including weekly updates, team line up with squad formation (X, Y coordinates), betting odds from up to 10 providers, detailed match events (goal types, corner, possession, fouls, etc.) for 10,000+ matches. For non-commercial use only.

URL: \url{https://www.kaggle.com/hugomathien/soccer}

\item \emph{Football Analytics} ---  This dataset contains European footbal team stats Only teams of Premier League, Ligue 1, Bundesliga, Serie A and La Liga are listed. The auxiliary datasets contain 2021--2022 Football Player Stats and 2021--2022 Football Team Stats.

URL: \url{https://www.kaggle.com/datasets/vivovinco/football-analytics}

\item \emph{Football DataSet} --- 96,000+ matches with detailed minute-by-minute history of the single game and players names, goals, yellow/red cards, penalties, var, penalties missed, etc. Season 2021--2022 included. 18 European Leagues from 10 countries with their lead championship.

URL: \url{https://www.kaggle.com/datasets/bastekforever/complete-football-data-89000-matches-18-leagues}

\item The Football Dataset from the Football Computer Vision project~\cite{Roboflow2023}.

URL: \url{https://universe.roboflow.com/football-detect/football-xrbge}

\item RapidAPI lists several football APIs that provide football (soccer) data for developers.

URL: \url{https://rapidapi.com/collection/football-soccer-apis}

\end{enumerate}

\section{Conclusion}

In this, necessarily incomplete, compendium we have provided some citations and/or links to datasets that we, and/or our collaborators, consider interesting and useful. We hope to update this list in future editions.

\bibliographystyle{plain}

\end{document}